\documentclass[12pt,letter]{article}

\usepackage[utf8]{inputenc}
\usepackage[english]{babel}
\usepackage{natbib}
\usepackage{authblk}
\usepackage{amssymb}
\usepackage[table]{xcolor}

\usepackage{hyperref}

\usepackage{tabularx}
\usepackage{graphicx}
\usepackage{booktabs}
\usepackage[para,online,flushleft]{threeparttable}
\usepackage{makecell}
\usepackage{comment}
\usepackage{paralist}
\usepackage[margin=0.5in]{geometry}

\usepackage{enumitem}

\newcommand{\blue}[1]{\textcolor{blue}{#1}}
\renewcommand{\blue}[1]{#1}

\definecolor{newcolor}{rgb}{.8,.349,.1}

\hypersetup{
    colorlinks  = true,
    urlcolor    = black!50!blue,
    linkcolor   = black!50!blue,
    citecolor   = black!50!blue,
}

\title{Towards long-tailed, multi-label disease classification from chest X-ray: Overview of the CXR-LT challenge}

\author[1]{\normalsize Gregory {Holste}}
\author[2]{Yiliang {Zhou}}
\author[1]{Song {Wang}}
\author[1]{Ajay {Jaiswal}}
\author[2]{Mingquan {Lin}}
\author[3]{Sherry {Zhuge}}
\author[4]{Yuzhe {Yang}}
\author[5]{Dongkyun {Kim}}  
\author[6]{Trong-Hieu {Nguyen-Mau}}  
\author[6]{Minh-Triet {Tran}}  
\author[7]{Jaehyup {Jeong}}  
\author[8]{Wongi {Park}}  
\author[8]{Jongbin {Ryu}}  
\author[9]{Feng {Hong}}  
\author[10]{Arsh {Verma}}  
\author[11]{Yosuke {Yamagishi}}  
\author[12]{Changhyun {Kim}}  
\author[13]{Hyeryeong {Seo}}  
\author[14]{Myungjoo {Kang}}  
\author[15,16,17]{Leo Anthony {Celi}}
\author[18]{Zhiyong {Lu}}
\author[19]{Ronald M. {Summers}}
\author[20]{George {Shih}}
\author[1]{Zhangyang {Wang}$^*$}
\author[2]{Yifan {Peng}\footnote{Corresponding authors. Email: \href{mailto:yip4002@med.cornell.edu}{yip4002@med.cornell.edu}, \href{mailto:atlaswang@utexas.edu}{atlaswang@utexas.edu}.}}

\affil[1]{\footnotesize Department of Electrical and Computer Engineering, The University of Texas at Austin, 78712, Austin, TX USA}
\affil[2]{Department of Population Health Sciences, Weill Cornell Medicine, 10065, New York, NY USA}
\affil[3]{School of Information Systems, Carnegie Mellon University, 15213, Pittsburgh, PA USA}
\affil[4]{Department of Electrical Engineering and Computer Science, Massachussetts Institute of Technology, 02139, Cambridge, MA USA}
\affil[5]{School of Computer Science, Carnegie Mellon University, 15213, Pittsburgh, PA USA}
\affil[6]{University of Science, VNU-HCM, 70000, Ho Chi Minh City, Vietnam}
\affil[7]{KT Research \& Development Center, KT Corporation, 06763, Seoul, South Korea}
\affil[8]{Department of Software and Computer Engineering, Ajou University, 16499, Suwon, South Korea}
\affil[9]{Cooperative Medianet Innovation Center, Shanghai Jiao Tong University, 200240, Shanghai, China}
\affil[10]{Wadhwani Institute for Artificial Intelligence, 400079, Mumbai, India}
\affil[11]{Division of Radiology and Biomedical Engineering, Graduate School of Medicine, The University of Tokyo, 113-0033, Tokyo, Japan}
\affil[12]{BioMedical AI Team, AIX Future R\&D Center, SK Telecom, 04539, Seoul, South Korea}
\affil[13]{Interdisciplinary Program in AI (IPAI), Seoul National University, 02504, Seoul, South Korea}
\affil[14]{Department of Mathematical Sciences, Seoul National University, 02504, Seoul, South Korea}
\affil[15]{Laboratory for Computational Physiology, Massachusetts Institute of Technology, 02139, Cambridge, MA USA}
\affil[16]{Division of Pulmonary, Critical Care and Sleep Medicine, Beth Israel Deaconess Medical Center, 02215, Boston, MA USA}
\affil[17]{Department of Biostatistics, Harvard T.H. Chan School of Public Health, 02115, Boston, MA USA}
\affil[18]{National Center for Biotechnology Information, National Library of Medicine, 20894, Bethesda, MD USA}
\affil[19]{Clinical Center, National Institutes of Health, 20892, Bethesda, MD USA}
\affil[20]{Department of Radiology, Weill Cornell Medicine, 10065, New York, NY USA}

\date{}

\begin{document}

\maketitle

\vspace*{-15mm}
\begin{abstract}
	Many real-world image recognition problems, such as diagnostic medical imaging exams, are ``long-tailed" -- there are a few common findings followed by many more relatively rare conditions. In chest radiography, diagnosis is both a \textit{long-tailed} and \textit{multi-label} problem, as patients often present with multiple findings simultaneously. While researchers have begun to study the problem of long-tailed learning in medical image recognition, few have studied the interaction of label imbalance and label co-occurrence posed by long-tailed, multi-label disease classification. To engage with the research community on this emerging topic, we conducted an open challenge, \textbf{CXR-LT}, on long-tailed, multi-label thorax disease classification from chest X-rays (CXRs). We publicly release a large-scale benchmark dataset of over 350,000 CXRs, each labeled with at least one of 26 clinical findings following a long-tailed distribution. We synthesize common themes of top-performing solutions, providing practical recommendations for long-tailed, multi-label medical image classification. Finally, we use these insights to propose a path forward involving vision-language foundation models for few- and zero-shot disease classification.\\


\end{abstract}

\section{Introduction}

Like many diagnostic medical exams, chest X-rays (CXRs) yield a long-tailed distribution of clinical findings. This means that while a small subset of diseases are routinely observed, the majority are quite rare \citep{zhou2021review}. This long-tailed distribution challenges conventional deep learning methods, as they tend to favor common classes and often overlook the infrequent yet crucial classes. In response, several methods \citep{zhang2023deep} have been proposed lately with a focus on addressing label imbalance in long-tailed medical image recognition tasks \citep{zhang2021mbnm,ju2021relational,ju2022flexible,yang2022proco}. Of note, diagnosing from CXRs is not only a long-tailed problem, but also \textit{multi-label}, since patients often present with multiple disease findings simultaneously. Despite this, only a limited number of studies have incorporated knowledge of label co-occurrence into their learning process \citep{chen2020label,wang2023bb,chen2019deep}.

Owing to the fact that most large-scale image classification benchmarks feature single-label images with a predominately balanced label distribution, we establish a new benchmark for long-tailed, multi-label medical image classification. Specifically, we expanded the MIMIC-CXR \citep{johnson2019mimic} dataset by increasing the set of target disease findings from 14 to 26. This is achieved by introducing 12 new disease findings by parsing the radiology reports associated with each CXR study. 

In our effort to engage with the community on this emerging interdisciplinary topic, we have released the data and launched the \textbf{CXR-LT} challenge on long-tailed, multi-label thorax disease classification on CXRs. In this paper, we summarize the CXR-LT challenge, consolidate key insights from top-performing solutions, and offer practical perspetive for advancing long-tailed, multi-label medical image classification. Finally, we use our findings to suggest a path forward toward few- and zero-shot disease classification in the long-tailed, multi-label setting by leveraging multimodal foundation models.

Our contributions can be summarized as follows:
\begin{enumerate}
\item We have publicly released a large multi-label, long-tailed CXR dataset containing 377,110 images. Each image is labeled with one or multiple labels from a set of 26 disease findings. In \blue{addition}, we have provided a ``gold standard" subset encompassing human-annotated consensus labels.
\item We conducted \textbf{CXR-LT}, a \blue{challenge} for long-tailed, multi-label thorax disease classification on CXRs. We summarize insights from top-performing teams and offer practical recommendations for advancing long-tailed, multi-label medical image classification.
\item Based on the insights from CXR-LT, we propose a methodological path forward for few- and zero-shot generalization to unseen disease findings via multimodal foundation models.
\end{enumerate}

\section{Methods}

\subsection{Dataset curation}

In this section, we detail the data curation process of two datasets: (i) the CXR-LT dataset used in the challenge, and (ii) a manually annotated ``gold standard" test set used for additional evaluation of top-performing solutions after the conclusion of the challenge.

\subsubsection{CXR-LT dataset}
\label{sec:data}

The CXR-LT challenge dataset\footnote{\url{https://physionet.org/content/cxr-lt-iccv-workshop-cvamd/1.1.0/}} was created by extending the label set of the MIMIC-CXR dataset\footnote{\url{https://physionet.org/content/mimic-cxr/2.0.0/}} \citep{johnson2019mimic}, resulting in a more challenging, long-tailed label distribution. \blue{MIMIC-CXR is a large-scale, publicly available dataset of de-identified CXRs and radiology reports. The dataset contains a total of 377,110 frontal and lateral CXR images acquired from 227,835 studies conducted during routine clinical practice at the Beth Israel Deaconess Medical Center (Boston, Massachusetts, USA) emergency department from 2011-2016.}

Following \cite{holste2022long}, the radiology reports associated with each CXR study were parsed via RadText \citep{wang2022radtext}, a radiology text analysis tool, to extract the presence status of \textit{12 new rare disease findings}:
\begin{inparaenum}[(1)]
\item Calcification of the Aorta,
\item Emphysema,
\item Fibrosis,
\item Hernia,
\item Infiltration,
\item Mass,
\item Nodule,
\item Pleural Thickening,
\item Pneumomediastinum,
\item Pneumoperitoneum,
\item Subcutaneous Emphysema, and
\item Tortuous Aorta.
\end{inparaenum}
\blue{These particular findings were selected by (i) identifying diseases found in the NIH \blue{ChestX-Ray} dataset \citep{wang2017chestx} labels that were not present in the MIMIC-CXR labels and (ii) discussing with radiologists which findings might be important to include that were not captured by existing public CXR datasets. The latter point led to the inclusion of Calcification of the Aorta and Tortuous Aorta (cardiac findings) as well as Pneumomediastinum, Pneumoperitoneum, and Subcutaneous Emphysema (trapped air in undesirable cavities or under the skin).}

\begin{figure*}
\centering
\includegraphics[scale=0.67]{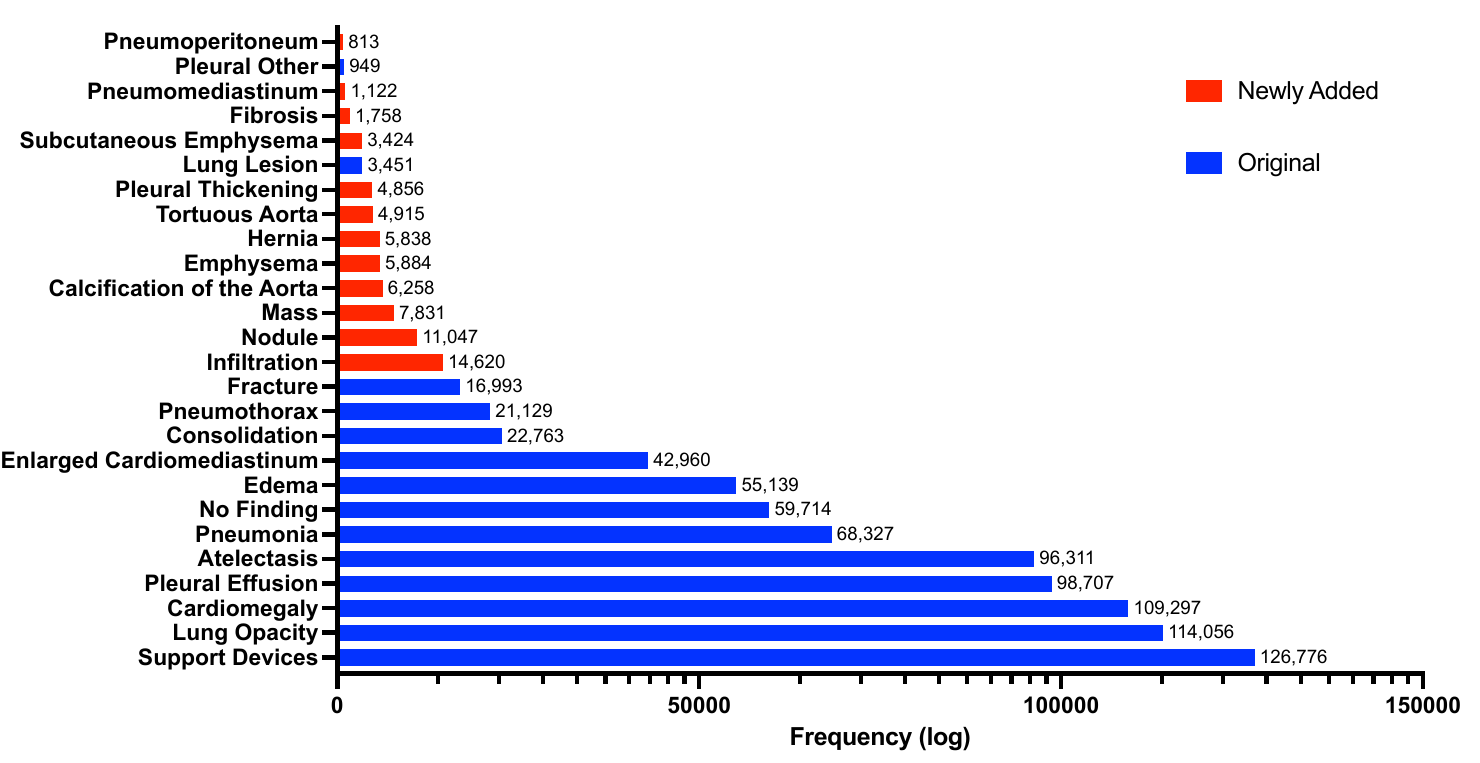}
\caption{Long-tailed distribution of the CXR-LT 2023 challenge dataset. The dataset was formed by extending the MIMIC-CXR \citep{johnson2019mimic} benchmark to include 12 new clinical findings (red) by parsing radiology reports.}
\label{fig1}
\end{figure*}

The resulting dataset consisted of 377,110 CXRs, each labeled with at least one of 26 disease findings following a long-tailed distribution (Fig.~\ref{fig1}). Though MIMIC-CXR contained the images and text reports needed for additional labeling, we used images from the MIMIC-CXR-JPG dataset \citep{johnson2019mimicjpg} in this \blue{challenge} since the preprocessed JPEG images ($\sim$600 GB) would be more accessible to participants than the raw DICOM data ($\sim$4.7 TB) provided in MIMIC-CXR.\footnote{\url{https://physionet.org/content/mimic-cxr-jpg/2.0.0/}}
\blue{Finally}, the dataset was randomly split into training (70\%), development (10\%), and test sets (20\%) at the patient level to avoid label leakage. \blue{Challenge} participants would have access to all images, but only have access to labels for the training set. 

\begin{figure}[!ht]
	\centering
	\includegraphics[scale=0.7]{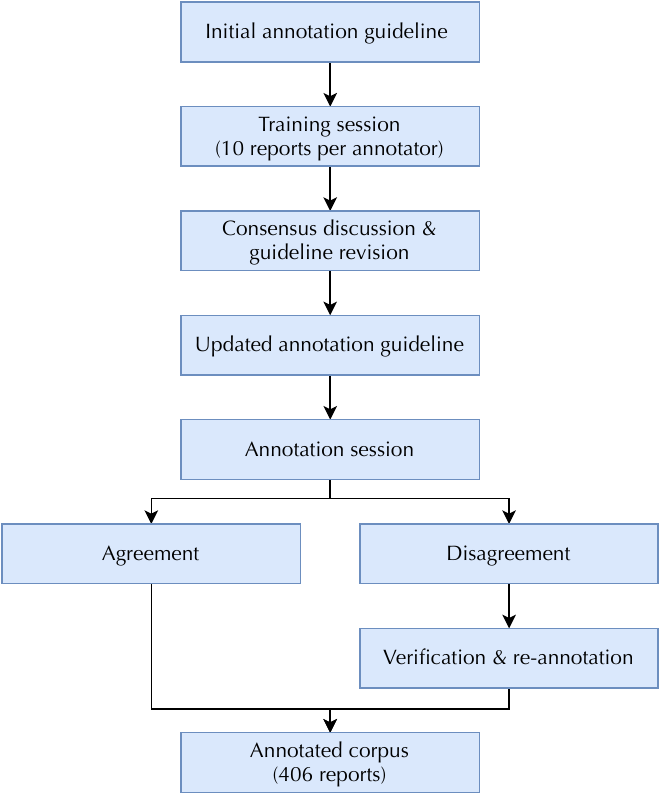}
	\caption{Flowchart describing CXR-LT gold standard dataset annotation.}
	\label{fig:gold-flowchart}
\end{figure}

\subsubsection{Gold standard test set}
\label{sec:gold-data}

While the CXR-LT dataset is large and challenging due to heavy label imbalance and label co-occurrence, it inevitably suffers from label noise much like other datasets with automatically text-mined labels \citep{abdalla2023hurdles}. To remedy this, we aimed to construct a ``gold standard" set, derived from the challenge test set, with labels that were manually annotated after analyzing the radiology reports. This smaller, but higher quality, dataset would then be used as an auxiliary test set to perform additional evaluation of the top-performing CXR-LT solutions \blue{after the conclusion of the challenge}.

To build a gold standard set for evaluation, \blue{we first curated a sample of test set reports with at least two positive disease findings as determined by RadText. This was necessary to ensure a sufficient number of positive examples for tail classes; given the extreme rarity of certain findings (as low as 0.2\% prevalence), a random sample of several hundred reports may not even yield a single instance of several rare findings, prohibiting proper performance evaluation. A subset of 451 such reports were manually annotated by six human readers, who marked the presence or absence of the 26 disease findings in each radiology report. Before manual labeling}, all reports were preprocessed through RadText \citep{wang2022radtext} to identify and highlight all relevant disease mentions in the text in order to ease the annotation process. Each annotator was then provided with the reports and a list of synonyms for each of the 26 findings. Annotators were asked to select all disease findings that were \blue{\textit{conclusively}} affirmed positive in the report. Following MIMIC-CXR, annotators could select ``No Finding" if no other findings (except ``Support Devices") were present.

Before annotation, a training session was held to align the standards among annotators where each annotator practiced by labeling 10 reports. Any disagreements in this phase were discussed until consensus was reached, leading to the formulation of a shared annotation guideline (Fig.~\ref{fig:gold-flowchart}). Following the training session, the official annotation process consisted of two rounds: the first round covering 200 reports and the second round covering 251 reports. After each round, individual disease-level disagreements between annotators on a given report were compiled and adjudicated by a third annotator. For the first round, the overall agreement rate was 93.2\% and the Cohen's Kappa coefficient was 0.795; for the second round, the agreement rate was 94.9\% with a Cohen's kappa of 0.778. After removing reports that were not annotated by at least two readers, the CXR-LT gold standard set consisted of 406 cases. The resulting label distribution of the gold standard set can be found in Supplementary Fig. 1.

\begin{figure*}[!ht]
	\centering
	\includegraphics[scale=0.7]{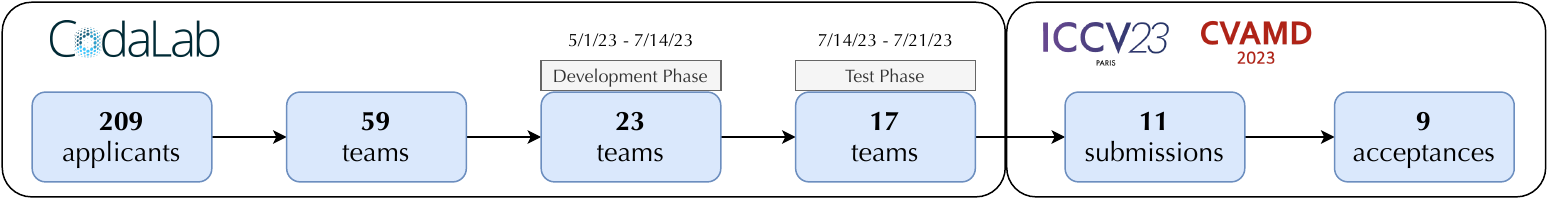}
	\caption{Flowchart describing CXR-LT challenge participation. Over 200 teams applied to participate in the challenge on CodaLab, and 59 teams met registration requirements. Of the 17 teams that participated in the Test Phase, 11 submitted their written solutions for presentation at the ICCV CVAMD 2023 workshop. The top 9 of these submissions were accepted to the workshop and are described in this paper.}
	\label{fig:cxr-lt_flowchart}
\end{figure*}

\subsection{CXR-LT challenge task}

The CXR-LT challenge was formulated as a 26-way multi-label classification problem. Given a CXR, participants were tasked with detecting all disease findings present. If no findings were present, participants could predict ``No Finding", with the exception that ``No Finding" can co-occur with ``Support Devices" as this is not a clinically meaningful \textit{diagnostic} finding. Since this is a multi-label classification problem with severe label imbalance, the primary evaluation metric was mean average precision (mAP), specifically, the ``macro-averaged" AP across the 26 classes. While area under the receiver operating characteristic curve (AUROC) is a standard metric employed for related datasets \citep{wang2017chestx,seyyed2021chexclusion}, AUROC can be heavily inflated in the presence of class imbalance \citep{fernandez2018learning,davis2006relationship}. Instead, mAP is more appropriate for the long-tailed, multi-label setting since it measures performance across decision thresholds and does not degrade under class imbalance \citep{rethmeier2022long}. For thoroughness, mean AUROC (mAUROC) and mean F1 score -- using a decision threshold of 0.5 for each class -- were also calculated.

The \blue{challenge} was conducted on CodaLab \citep{codalab_competitions_JMLR}. Any registered CodaLab user could apply to participate, but since this competition used MIMIC-CXR-JPG data \citep{johnson2019mimicjpg}, which requires credentialing and training through PhysioNet \citep{PhysioNet}, participants were required to submit proof of PhysioNet credentials to enter. During the Development Phase, registered participants downloaded the labeled training set and (unlabeled) development set, for which they would generate a comma-separated values (CSV) file with predictions to upload. Submissions were then evaluated on the held-out development set and results were updated to a live, public leaderboard. During the Test Phase, test set images (without labels) were released. Participants were asked to submit CSV files with predictions on the much larger, held-out test set and were only given a maximum of 5 successful attempts. For this phase, the leaderboard was kept hidden and the single best-scoring submission (by mAP) by each team was retained. The final Test Phase leaderboard was used to rank participants, primarily by mAP, then by mAUROC in the event of ties.

\section{Results}

\subsection{CXR-LT challenge participation}

The CXR-LT challenge received 209 team applications on CodaLab,\footnote{\url{https://codalab.lisn.upsaclay.fr/competitions/12599}} of which 59 were approved after providing proof of credentialed access to MIMIC-CXR-JPG \citep{johnson2019mimicjpg}. During the Development Phase, 23 teams participated, contributing a total of 525 unique submissions to the public leaderboard. Ultimately, 17 teams participated in the final Test Phase, and 11 of these teams submitted papers describing their challenge solution to the ICCV CVAMD 2023 workshop.\footnote{\url{https://cvamd2023.github.io/}} The 9 accepted workshop papers, representing the top-performing teams in the CXR-LT challenge, were used for study in this paper (Fig. \ref{fig:cxr-lt_flowchart}).

\subsection{Methods of top-performing teams}

A summary of top-performing solutions can be found in Table \ref{tab:overview}, including Test Phase rank, image resolution, backbone architecture used, and other methodological characteristics. Though each solution is described in the paragraphs below, please refer to the paper in each subsection title for full details.

\subsubsection{T1~\citep{Kim_2023_ICCVa}}
This team used a two-stage framework that aggregated features across views (e.g., frontal and lateral CXRs). In the first stage, a ConvNeXt-S model \citep{liu2022convnet} model was pretrained with Noisy Student \citep{xie2020self} self-training on the external CXR datasets NIH \blue{ChestX-Ray} \citep{wang2017chestx}, CheXpert 
\citep{irvin2019chexpert}, and VinDrCXR \citep{nguyen2022vindr}. Using this backbone as a frozen feature extractor, a Transformer then aggregated multi-view features in a given study. T1 used the ML-Decoder \citep{ridnik2023ml} classification head, which represents the labels as text and performs cross-attention over image (CXR) and text (label) features. Finally, this team utilized a weighted asymmetric loss \citep{ridnik2021asymmetric} to combat the inter-class imbalance caused by the long-tailed distribution and intra-class imbalance caused by the dominance of negative labels in multi-label classification.


\subsubsection{T2~\citep{Nguyen-Mau_2023_ICCV}}

This team utilized augmentation, ensemble, and \blue{re-weighting} methods for imbalanced multi-label classification. Specifically, they used an ensemble of EfficientNetV2-S \citep{tan2021efficientnetv2} and ConvNeXt-S \citep{liu2022convnet} models. Of note, the team made use of heavy ``mosaic" augmentation \citep{bochkovskiy2020yolov4}, randomly tiling four CXRs into a single image and using the union of their label sets as ground truth. They used a weighted focal loss \citep{lin2017focal} to handle imbalance, then test-time augmentation and a multi-level ensemble across both model architectures and individual models obtained by stratified cross-validation to improve generalization.



\begin{table*}[!ht]
	\caption{\label{tab:overview}Overview of top-performing CXR-LT challenge solutions. ENS = ensemble; RW = loss \blue{re-weighting}.}
	\centering
	\begin{threeparttable}
		\footnotesize
		\begin{tabularx}{\textwidth}{lll>{\raggedright\arraybackslash}p{2.6cm}cc>{\raggedright\arraybackslash}p{2.6cm}X}
			\toprule
			Team & Rank & \makecell[l]{Image\\Resolution} & Backbone & ENS & RW & Pretraining & Notes \\
			\midrule
			T1 & 1 & 1024 & ConvNeXt-S & & \checkmark & ImageNet $\rightarrow$ CheXpert, NIH, VinDr & Two-stage training; cross-view Transformer; ML-Decoder classifier (label as text)\\
			\arrayrulecolor{black!30}\midrule
			T2 & 2 & 512, 768 & EfficientNetV2-S, ConvNeXt-S & \checkmark & \checkmark & ImageNet & Heavy mosaic augmentation \\
			\midrule
			T3 & 3 & 448 & ConvNeXt-B & \checkmark & \checkmark & ImageNet21K $\rightarrow$ NIH & Ensemble of ``head" and ``tail" experts \\
			\midrule
			T4  & 4 & 384 & ConvNeXt-B & & \checkmark & ImageNet & Custom robust asymmetric loss (RAL) \\
			\midrule
			T5 & 5 & 512 & ResNet50* & \checkmark & \checkmark & ImageNet* & \makecell[l]{Vision-language modeling (label as text);\\co-train on NIH, CheXpert} \\
			\midrule
			T6  & 6 & 448 & ResNeXt101, DenseNet161 & \checkmark & & ImageNet $\rightarrow$ CheXpert, NIH, PadChest & Used synthetic data to augment tail classes \\
			\midrule
			T7  & 7 & 224$-$512 & EfficientNetV2-S & \checkmark & & ImageNet, ImageNet21k & Three-stage training with increasing resolution \\
			\midrule
			T8 & 8 & 448 & TResNet50 & \checkmark & & ImageNet & \makecell[l]{Heavy CutMix-like augmentation;\\feature pyramid with deep supervision} \\
			\midrule
			T9 & 11 & 1024 & ResNet101 & \checkmark & & ImageNet & RIDE mixture of experts; LSE pooling; label as text/graph with cross-modal attention \\
			\arrayrulecolor{black!}\bottomrule
		\end{tabularx}
		\begin{tablenotes}
			*T5 additionally used a Transformer text encoder pretrained on PubMedBERT \citep{pubmedbert} and Clinical-T5 \citep{lehman2023clinical}.
		\end{tablenotes}
	\end{threeparttable}
\end{table*}

\subsubsection{T3~\citep{Jeong_2023_ICCV}}

This team proposed an ensemble method based on ConvNeXt-B \citep{liu2022convnet} with the CSRA classifier \citep{zhu2021residual}. After pretraining on the NIH \blue{ChestX-Ray} \citep{wang2017chestx} dataset, T3 trained three separate models, respectively, on CXR-LT data only from ``head" classes, ``tail" classes, and all classes; an average of these three models formed the final output. Each model utilized a weighted cross-entropy loss and the Lion optimizer \citep{chen2023symbolic}.


\subsubsection{T4~\citep{Park_2023_ICCV}}
This team proposed a novel robust asymmetric loss (RAL) for multi-label long-tailed classification. RAL improves upon the popular focal loss \citep{lin2017focal} by including a Hill loss term \citep{zhang2021simple}, which mitigates sensitivity to the negative term of the original focal loss. The team used an ImageNet-pretrained ConvNeXt-B \citep{liu2022convnet} with the proposed RAL loss and data augmentation following \citep{azizi2021big} and \citep{chen2019multi}.



\subsubsection{T5~\citep{hong2023bag}}
This team used a vision-language modeling approach leveraging large pre-trained models. The authors utilized an ImageNet-pretrained ResNet50 \citep{he2016deep} and text encoder pre-trained on PubMedBERT \citep{pubmedbert} and Clinical-T5 \citep{lehman2023clinical} to extract features from images and label text, respectively. For multi-label classification, they employed a multi-label Transformer query network to aggregate image and text features. To handle imbalance, the team used class-specific loss \blue{re-weighting} informed by validation set performance. They also incorporated external training data (NIH \blue{ChestX-Ray}\citep{wang2017chestx} and CheXpert \citep{irvin2019chexpert}), used test-time augmentation, and performed ``class-wise" ensembling to improve generalization.


\subsubsection{T6~\citep{verma2023can}}
This team used domain-specific pretraining, ensembling, and synthetic data augmentation to improve performance. With an ImageNet-pretrained ResNeXt101 \citep{xie2017aggregated} and DenseNet101 \citep{huang2017densely}, the team further pretrained on the CXR benchmarks NIH \blue{ChestX-Ray} \citep{wang2017chestx}, CheXpert \citep{irvin2019chexpert}, and PadChest \citep{bustos2020padchest}. This team also used RoentGen \citep{chambon2022roentgen}, a multimodal generative model for synthesizing CXRs from natural language, to generate additional CXRs for tail classes in order to combat imbalance.  


\begin{table*}[!t]
\caption{\label{tab:results}Final \blue{Test Phase} results of the CXR-LT 2023 \blue{challenge}. Presented is average precision (AP) of each team's final model on all 26 classes evaluated on the test set. The best AP for a given class is highlighted in bold.}
\centering
\begin{tabular}{lcccccccccc}
	\toprule
	& T1 & T2 & T3 & T4 & T5 & T6 & T7 & T8 & T9 \\ \midrule
	Atelectasis & \textbf{0.622} & 0.609 & 0.611 & 0.607 & 0.606 & 0.610 & 0.602 & 0.595 & 0.546 \\
	Calcification of the Aorta & \textbf{0.162} & 0.140 & 0.145 & 0.143 & 0.135 & 0.109 & 0.130 & 0.116 & 0.111 \\
	Cardiomegaly & 0.661 & 0.652 & 0.652 & 0.648 & 0.653 & 0.652 & \textbf{0.668} & 0.640 & 0.581 \\
	Consolidation & \textbf{0.240} & 0.228 & 0.234 & 0.219 & 0.228 & 0.230 & 0.225 & 0.218 & 0.171 \\
	Edema & \textbf{0.563} & 0.553 & 0.559 & 0.556 & 0.554 & 0.557 & 0.551 & 0.545 & 0.497 \\
	Emphysema & \textbf{0.210} & 0.193 & 0.193 & 0.193 & 0.180 & 0.184 & 0.161 & 0.165 & 0.128 \\
	Enlarged Cardiomediastinum & 0.186 & 0.184 & 0.184 & 0.184 & \textbf{0.186} & 0.185 & 0.183 & 0.177 & 0.140 \\
	Fibrosis & \textbf{0.167} & 0.163 & 0.157 & 0.153 & 0.154 & 0.154 & 0.116 & 0.132 & 0.120 \\
	Fracture & \textbf{0.379} & 0.262 & 0.269 & 0.289 & 0.243 & 0.262 & 0.171 & 0.219 & 0.171 \\
	Hernia & 0.570 & \textbf{0.585} & 0.563 & 0.551 & 0.539 & 0.538 & 0.499 & 0.484 & 0.343 \\
	Infiltration & \textbf{0.063} & 0.057 & 0.060 & 0.060 & 0.057 & 0.058 & 0.056 & 0.055 & 0.049 \\
	Lung Lesion & 0.034 & \textbf{0.042} & 0.041 & 0.038 & 0.040 & 0.031 & 0.031 & 0.031 & 0.021 \\
	Lung Opacity & \textbf{0.617} & 0.597 & 0.603 & 0.597 & 0.594 & 0.598 & 0.590 & 0.584 & 0.529 \\
	Mass & \textbf{0.250} & 0.224 & 0.213 & 0.206 & 0.200 & 0.227 & 0.187 & 0.167 & 0.112 \\
	Nodule & \textbf{0.267} & 0.192 & 0.204 & 0.200 & 0.180 & 0.196 & 0.137 & 0.166 & 0.117 \\
	Pleural Effusion & \textbf{0.843} & 0.829 & 0.831 & 0.832 & 0.830 & 0.805 & 0.822 & 0.822 & 0.781 \\
	Pleural Other & \textbf{0.070} & 0.037 & 0.043 & 0.040 & 0.039 & 0.016 & 0.059 & 0.042 & 0.007 \\
	Pleural Thickening & \textbf{0.137} & 0.108 & 0.116 & 0.110 & 0.126 & 0.083 & 0.119 & 0.097 & 0.055 \\
	Pneumomediastinum & 0.332 & 0.384 & 0.376 & 0.339 & \textbf{0.387} & 0.284 & 0.326 & 0.308 & 0.096 \\
	Pneumonia & \textbf{0.312} & 0.305 & 0.311 & 0.309 & 0.304 & 0.308 & 0.292 & 0.294 & 0.258 \\
	Pneumoperitoneum & \textbf{0.324} & 0.316 & 0.261 & 0.283 & 0.303 & 0.237 & 0.262 & 0.235 & 0.155 \\
	Pneumothorax & \textbf{0.602} & 0.533 & 0.549 & 0.553 & 0.511 & 0.546 & 0.451 & 0.474 & 0.427 \\
	Subcutaneous Emphysema & \textbf{0.598} & 0.556 & 0.564 & 0.560 & 0.570 & 0.520 & 0.507 & 0.538 & 0.492 \\
	Support Devices & \textbf{0.918} & 0.906 & 0.916 & 0.913 & 0.910 & 0.910 & 0.894 & 0.903 & 0.887 \\
	Tortuous Aorta & \textbf{0.066} & 0.061 & 0.060 & 0.060 & 0.058 & 0.056 & 0.063 & 0.053 & 0.045 \\
	No Finding & 0.486 & 0.478 & \textbf{0.488} & 0.479 & 0.485 & 0.468 & 0.471 & 0.469 & 0.428 \\
	\midrule
	Mean & \textbf{0.372} & 0.354 & 0.354 & 0.351 & 0.349 & 0.339 & 0.330 & 0.328 & 0.279 \\ \bottomrule
\end{tabular}
\end{table*}

\subsubsection{T7~\citep{Yamagishi_2023_ICCV}}
This team used a multi-stage training scheme with ensembling and oversampling. For the first stage, an ImageNet21k-pretrained EfficientNetV2-S \citep{tan2021efficientnetv2} was trained on $224 \times 224$ resolution images. The weights from this model were then used to train on $320 \times 320$ and $384 \times 384$ images, then $512 \times 512$ images. An ensemble was then formed by averaging the predictions of four models trained on various resolutions, with some models leveraging oversampling of minority classes to mitigate class imbalance. Test-time augmentation and view-based post-processing were also used to boost performance.

\subsubsection{T8~\citep{Kim_2023_ICCVb}}
This  team utilized an ImageNet-pretrained TResNet50 \citep{ridnik2021tresnet} with heavy augmentation and ensembling. The authors made use of MixUp \citep{zhang2017mixup}, which linearly combines training images and their labels, and CutMix \citep{yun2019cutmix}, which ``cuts and pastes" regions from one training image onto another. They also used a ``feature pyramid" approach, extracting pooled features from four layers throughout the network and aggregating these multi-scale features.

\subsubsection{T9~\citep{Seo_2023_ICCV}}
This team built upon ML-GCN \citep{chen2019multi}, a framework for multi-label image classification, which uses GloVe \citep{pennington2014glove} to embed each label as a node within a graph capable of incorporating the co-occurrence patterns of labels. To combat the long-tailed distribution of classes, the authors trained a ResNet101 \citep{he2016deep} with class-balanced sampling and the Routing DIverse Experts (RIDE) method \citep{wang2020long} to diversify the members of their ensemble \citep{zhang2023deep}. They also used log-sum-exp (LSE) pooling \citep{pinheiro2015image} and a Transformer encoder to attend over image and text features.

\subsection{\blue{CXR-LT challenge results}}

\subsubsection{CXR-LT Test Phase results}
Detailed Test Phase results of 9 top-performing CXR-LT teams can be found in Table \ref{tab:results}. T1, the 1\textsuperscript{st}-placed team, reached an mAP of 0.372, considerably outperforming the 2\textsuperscript{nd}-5\textsuperscript{th}-placed teams, who performed similarly with mAP ranging from 0.349 to 0.354; further, T1 achieved best performance on 20 out of 26 classes. Maximum per-class AP ranged widely from 0.063 (Infiltration) to 0.918 (Support Devices), \blue{likely} owing to the challenges posed by label imbalance and noise. \blue{Interestingly, T1 demonstrated outstanding performance particularly on the Fracture (0.379 mAP vs. next-best 0.289) and Nodule (0.267 mAP vs. next-best 0.204) classes. Since T1 was the only team to explicitly fuse multi-view information, it is conceivable that this learned aggregation of frontal- and lateral-view information aided performance particularly for findings like Fracture and Nodule, which may better be resolved by multiple views.} Additional Test Phase results by AUROC can be found in Supplementary Table 1. 

\begin{table}[!ht]
	\caption{\label{tab:lt-results}\blue{Long-tailed classification performance on “head”, “medium”, and “tail” classes by average mAP within each category. These categories were determined by relative frequency of each class in the training set (denoted in parentheses). The rightmost column denotes the average of head, medium, and tail mAP. The best mAP in each column appears in bold.}}
	\centering
	
	\begin{tabular}{@{}cccccc@{}}
		\toprule
		& Overall & \makecell[c]{Head\\($>$10\%)} & \makecell[c]{Medium\\(1-10\%)} & \makecell[c]{Tail\\($<$1\%)} & Avg \\ \midrule
		T1 & \textbf{0.372} & \textbf{0.499} & \textbf{0.246} & 0.242 & \textbf{0.329}  \\
		T2 & 0.354 & 0.482 & 0.226 & 0.227 & 0.311  \\
		T3 & 0.354 & 0.477 & 0.226 & \textbf{0.246} & 0.316  \\
		T4 & 0.351 & 0.480 & 0.221 & 0.220 & 0.307 \\
		T5 & 0.349 & 0.474 & 0.218 & 0.243 & 0.312 \\
		T6 & 0.339 & 0.476 & 0.210 & 0.179 & 0.288 \\
		T7 & 0.330 & 0.460 & 0.195 & 0.216 & 0.290 \\
		T8 & 0.328 & 0.461 & 0.195 & 0.195 & 0.284 \\
		T9 & 0.279 & 0.420 & 0.154 & 0.086 & 0.220 \\ \bottomrule
	\end{tabular}
\end{table}

\begin{table}[ht]
\caption{\label{tab:annot-diff}\blue{Comparison of disease prevalence with automated text-mined labeling vs. manual human annotation of the 406 radiology reports in our gold standard test set.}}
\centering
\begin{tabular}{@{}l@{~~}ccr@{}}
	\toprule
	& Automated & Human & \makecell[l]{$\Delta$} \\ \midrule
	Atelectasis & 0.488 & 0.308 & $-$37\% \\
	Calcification of the Aorta & 0.062 & 0.116 & $+$88\% \\
	Cardiomegaly & 0.448 & 0.392 & $-$13\% \\
	Consolidation & 0.214 & 0.182 & $-$15\% \\
	Edema & 0.313 & 0.249 & $-$20\% \\
	Emphysema & 0.113 & 0.071 & $-$37\% \\
	Enlarged Cardiomediastinum & 0.313 & 0.291 & $-$7\% \\
	Fibrosis & 0.081 & 0.054 & $-$33\% \\
	Fracture & 0.133 & 0.118 & $-$11\% \\
	Hernia & 0.074 & 0.047 & $-$37\% \\
	Infiltration & 0.113 & 0.03 & $-$74\% \\
	Lung Lesion & 0.076 & 0.015 & $-$81\% \\
	Lung Opacity & 0.571 & 0.49 & $-$14\% \\
	Mass & 0.096 & 0.049 & $-$49\% \\
	No Finding & 0.091 & 0.091 & $+$0\% \\
	Nodule & 0.096 & 0.081 & $-$15\% \\
	Pleural Effusion & 0.562 & 0.451 & $-$20\% \\
	Pleural Other & 0.052 & 0.047 & $-$10\% \\
	Pleural Thickening & 0.052 & 0.054 & $+$5\% \\
	Pneumomediastinum & 0.099 & 0.086 & $-$12\% \\
	Pneumonia & 0.283 & 0.054 & $-$81\% \\
	Pneumoperitoneum & 0.084 & 0.059 & $-$29\% \\
	Pneumothorax & 0.214 & 0.121 & $-$44\% \\
	Subcutaneous Emphysema & 0.108 & 0.103 & $-$4\% \\
	Support Devices & 0.564 & 0.544 & $-$4\% \\
	Tortuous Aorta & 0.059 & 0.081 & $+$38\% \\ \bottomrule
\end{tabular}
\end{table}

\subsubsection{\blue{Long-tailed classification performance}}
\blue{To examine predictive performance by label frequency, we split the 26 target classes into ``head" ($>$10\%), ``medium" (1-10\%), and ``tail" ($<$1\%) categories based on prevalence in the training set. Category-wise mAP is presented in Table \ref{tab:lt-results}, as well as a ``category-wise average" of head, medium, and tail mAP. We first observe that T1 considerably outperformed all other teams, particularly on the more common head and medium classes, though other teams performed comparably or slightly better (T3 and T5) on tail classes. We also find that T1-T5, on average, outperformed the remaining four teams on tail classes much more so than head and medium classes. Critically, T1-T5 were the only five teams to use loss re-weighting, which appears to have provided clear benefits in modeling rare diseases.}

\subsection{\blue{Evaluation on a gold standard test set}}
\blue{As described in Section \ref{sec:gold-data}, a ``gold standard" test set of 406 newly labeled CXRs was curated for additional evaluation on a small subset of the challenge test set with higher-quality labels. This section describes differences in human vs. automated annotation patterns and predictive performance with human vs. automated labels in this gold standard set.}

\subsubsection{\blue{Differences in human vs. text-mined annotation}}
\label{lt-diff}
\blue{To form a head-to-head comparison of disease annotation patterns of human readers vs. automatic text mining tools, we examined the distribution of manual and automated labels on the 406 gold standard reports (Table \ref{tab:annot-diff}). Overall, we find that automatic labeling produced a higher volume of positive findings per image (median: 5, mean $\pm$ std: 5.4 $\pm$ 1.9) than manual annotation (median: 4, mean $\pm$ std: 4.2 $\pm$ 1.8). This is likely a consequence of the conservative labeling approach adopted by annotators after the initial training session. While we elected to only mark the presence of a finding if it was unambiguously affirmed positive in the report, text analysis tools like RadText do not necessarily adhere to the same rule. For example, potentially ambiguous phrases like “cannot rule out pneumonia” and “likely effusion” might be marked positive by a text analysis tool but not by human annotators.}

\blue{While manual labeling produced fewer positive findings in general, we find that certain diseases were substantially more and less likely to be marked positive by a human reader than RadText. For example, as seen in Table \ref{tab:annot-diff}, Pneumonia and Lung Lesion both saw an $>$80\% drop in prevalence with human labeling. On the other hand, Calcification of the Aorta (88\%) and Tortuous Aorta (38\%) – both newly added cardiac findings – were considerably more prevalent upon human annotation, suggesting a low recall for these aortic findings with RadText. Interestingly, there was no difference in the number of reports marked No Finding, indicating that it is perhaps straightforward for both humans and text analysis tools to recognize the absence of findings in a normal report.}

\blue{Since this is a multi-label classification problem, we also examine differences in co-occurrence behavior between findings in the text-mined vs. gold standard labels. We first computed a conditional co-occurrence probability matrix for each dataset, $C^{\textrm{test}}$ and $C^{\textrm{gold}}$, where entry $(i, j)$ represents the observed conditional probability that finding $j$ was present given finding $i$ was present: $C_{i,j} = P(j | i)$. We then computed $C^{\textrm{gold}} - C^{\textrm{test}}$, the difference in conditional co-occurrence probability between human vs. automated labels, visualized as a heatmap in Supplementary Fig. 2. We observe certain irregularities such as an $>$0.6 drop in $P(\textrm{Enlarged Cardiomediastinum} \ |\  \textrm{Pneumomediastinum})$ in the gold standard labels. Upon examination, we find that $P(\textrm{Enlarged Cardiomediastinum} \ |\  \textrm{Pneumomediastinum}) = 1$ in the text-mined labels, meaning every time Pneumomediastinum occurred, so did Enlarged Cardiomediastinum. This potentially reflects a shortcoming of text-mined labeling, where the presence of ``mediastinum” may be falsely interpreted as a positive assertion of both Pneumomediastinum and Enlarged Cardiomediastinum; in reality, these are two very different clinical findings that are unlikely to occur together.}

\begin{table*}[!ht]
	\caption{\label{tab:gold}Gold standard test set results from CXR-LT 2023 participants. Presented is average precision (AP) of each team's final model on all 26 classes evaluated on our human-annotated gold standard test set. The best AP for a given class is highlighted in bold.}
	\centering
	\begin{tabular}{lcccccccccc}
		\toprule
		& T1 & T2 & T3 & T4 & T5 & T6 & T7 & T8 & T9 \\ \midrule
		Atelectasis & 0.465 & 0.481 & 0.494 & 0.453 & \textbf{0.500} & 0.464 & 0.444 & 0.455 & 0.449 \\
		Calcification of the Aorta & 0.658 & 0.609 & \textbf{0.688} & 0.662 & 0.621 & 0.578 & 0.544 & 0.613 & 0.541 \\
		Cardiomegaly & 0.696 & 0.718 & 0.718 & 0.704 & 0.720 & 0.732 & \textbf{0.754} & 0.718 & 0.664 \\
		Consolidation & 0.415 & 0.449 & 0.471 & 0.474 & 0.436 & \textbf{0.476} & 0.411 & 0.426 & 0.406 \\
		Edema & 0.600 & 0.590 & \textbf{0.601} & 0.572 & 0.589 & 0.587 & 0.571 & 0.587 & 0.540 \\
		Emphysema & 0.298 & 0.356 & 0.362 & 0.309 & 0.359 & 0.388 & \textbf{0.394} & 0.279 & 0.292 \\
		Enlarged Cardiomediastinum & 0.351 & \textbf{0.370} & 0.349 & 0.349 & 0.337 & 0.341 & 0.328 & 0.314 & 0.337 \\
		Fibrosis & 0.417 & 0.491 & 0.460 & 0.458 & 0.487 & \textbf{0.497} & 0.397 & 0.437 & 0.428 \\
		Fracture & \textbf{0.583} & 0.455 & 0.535 & 0.494 & 0.524 & 0.501 & 0.385 & 0.366 & 0.389 \\
		Hernia & 0.759 & \textbf{0.808} & 0.804 & 0.766 & 0.722 & 0.723 & 0.714 & 0.708 & 0.514 \\
		Infiltration & 0.065 & 0.049 & 0.059 & 0.048 & 0.049 & 0.046 & 0.053 & 0.080 & \textbf{0.095} \\
		Lung Lesion & 0.028 & \textbf{0.071} & 0.033 & 0.042 & 0.030 & 0.032 & 0.066 & 0.040 & 0.044 \\
		Lung Opacity & 0.642 & 0.651 & \textbf{0.678} & 0.656 & 0.656 & 0.650 & 0.655 & 0.652 & 0.623 \\
		Mass & 0.410 & \textbf{0.477} & 0.389 & 0.376 & 0.369 & 0.412 & 0.321 & 0.363 & 0.128 \\
		Nodule & \textbf{0.435} & 0.325 & 0.296 & 0.300 & 0.272 & 0.359 & 0.362 & 0.257 & 0.212 \\
		Pleural Effusion & \textbf{0.856} & 0.835 & 0.836 & 0.842 & 0.831 & 0.808 & 0.822 & 0.837 & 0.804 \\
		Pleural Other & \textbf{0.385} & 0.212 & 0.224 & 0.259 & 0.230 & 0.121 & 0.249 & 0.245 & 0.138 \\
		Pleural Thickening & \textbf{0.386} & 0.249 & 0.249 & 0.204 & 0.244 & 0.190 & 0.251 & 0.210 & 0.132 \\
		Pneumomediastinum & 0.771 & 0.830 & 0.795 & 0.800 & 0.823 & 0.757 & \textbf{0.837} & 0.776 & 0.569 \\
		Pneumonia & 0.114 & \textbf{0.131} & 0.127 & 0.123 & 0.111 & 0.111 & 0.114 & 0.103 & 0.096 \\
		Pneumoperitoneum & 0.586 & 0.539 & \textbf{0.632} & 0.539 & 0.557 & 0.443 & 0.499 & 0.526 & 0.487 \\
		Pneumothorax & \textbf{0.675} & 0.649 & 0.701 & 0.641 & 0.598 & 0.663 & 0.562 & 0.568 & 0.587 \\
		Subcutaneous Emphysema & 0.845 & 0.825 & 0.852 & \textbf{0.887} & 0.874 & 0.777 & 0.774 & 0.813 & 0.837 \\
		Support Devices & 0.952 & 0.950 & 0.957 & \textbf{0.960} & 0.956 & 0.951 & 0.932 & 0.946 & 0.944 \\
		Tortuous Aorta & 0.362 & 0.329 & 0.322 & 0.309 & 0.336 & \textbf{0.386} & 0.265 & 0.299 & 0.280 \\
		No Finding & 0.731 & \textbf{0.841} & 0.834 & 0.729 & 0.852 & 0.828 & 0.815 & 0.807 & 0.766 \\
		\midrule
		Mean & \textbf{0.519} & 0.511 & 0.518 & 0.498 & 0.503 & 0.493 & 0.481 & 0.478 & 0.435 \\ \bottomrule
	\end{tabular}
\end{table*}

\begin{figure}[!ht]
\centering
\includegraphics[scale=0.7]{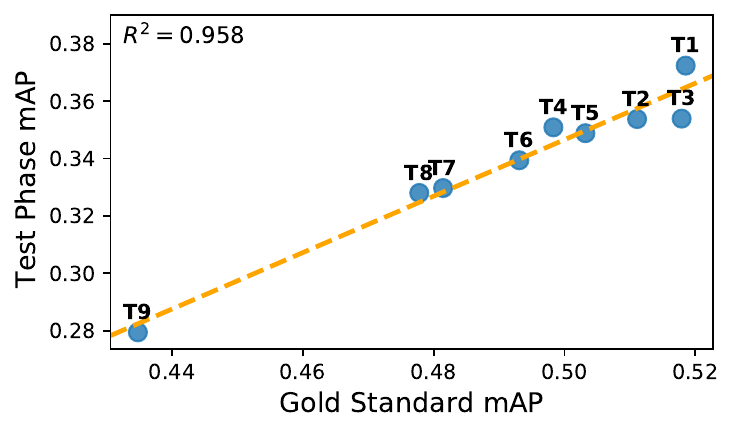}
\caption{Comparison of performance on CXR-LT Test Phase data (Section \ref{sec:data}) and gold standard test data (Section \ref{sec:gold-data}).}
\label{fig:test-vs-gold}
\end{figure}

\begin{figure*}[!ht]
\centering
\includegraphics[scale=0.7]{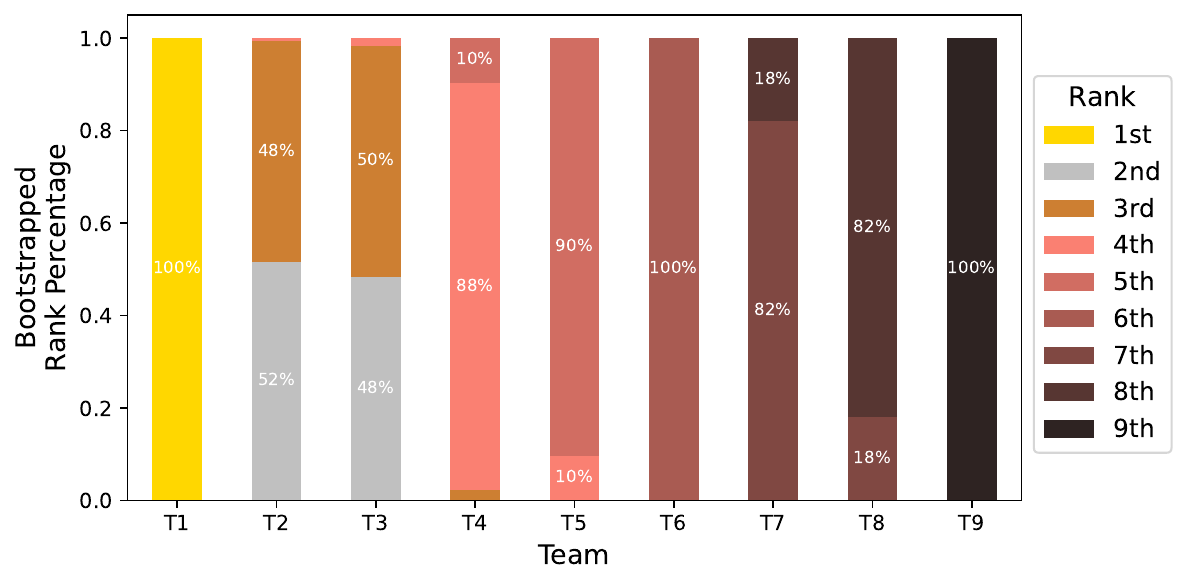}
\includegraphics[scale=0.7]{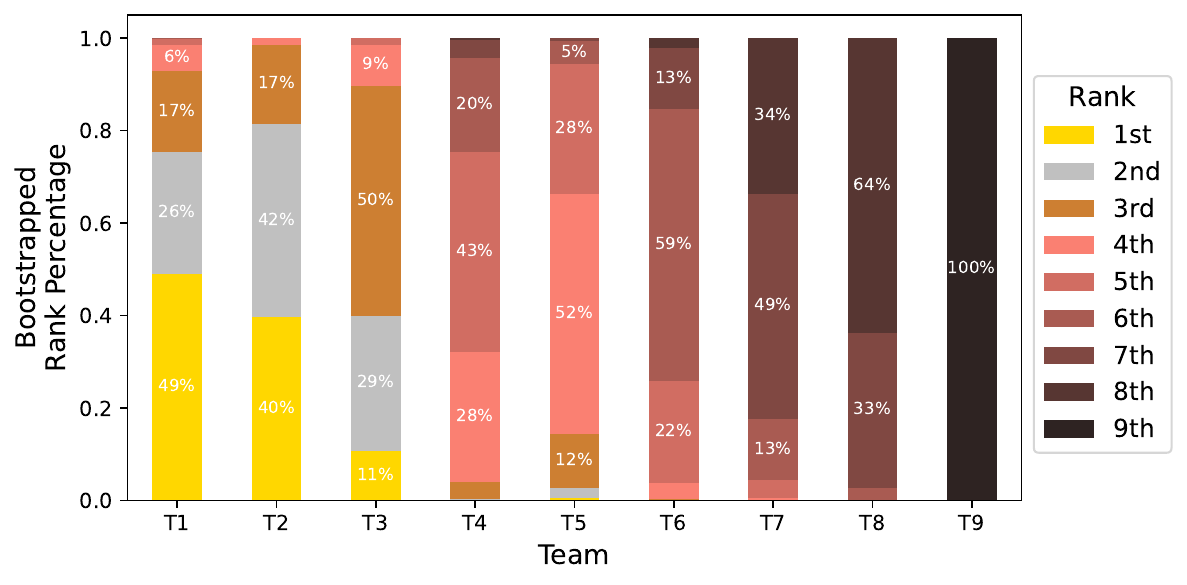}
\caption{\blue{Ranking stability analysis on CXR-LT Test Phase data (top) and gold standard test data (bottom). Simulated challenge rankings were computed via 500 bootstrap samples of each test set. The percentage of bootstrap trials for which each team achieved a given rank is presented.}}
\label{fig:ranking-stability}
\end{figure*}

\subsubsection{Gold standard test set \blue{results}}
Detailed results of top-performing teams on the gold standard test set can be found in Table \ref{tab:gold} \blue{(additional results by AUROC in Supplementary Table 2)}. AP values were generally higher in the gold standard set than the original challenge test set, with certain classes \blue{experiencing} large changes in performance -- for example, the maximum AP jumps from 0.162 to 0.688 for Calcification of the Aorta and from 0.598 to 0.887 for Subcutaneous Emphysema. \blue{However, these dramatic differences are to be expected when considering that the gold standard test set is not a representative subset of the challenge test set; for reasons outlined in Section \ref{sec:gold-data}, the gold standard set consisted of studies that were far more likely to contain many disease findings than the overall test set (median: 2, mean $\pm$ std: 2.4 $\pm$ 1.5). A direct comparison of class-wise performance -- specifically, the mean AP for each class across all 9 teams -- using automated text-mined vs. human labels in the 406 gold standard test cases can be found in Supplementary Table 3. Here, we observe that performance on ``gold standard" human labels vs. text-mined labels dropped for 19 out of 26 classes and experienced an average 12.3\% drop in AP. This can be attributed to the large label distribution shift outlined in Section \ref{lt-diff}, meaning models have been optimized on the particular noise patterns inherent in the text-mined training set labels.}

Despite \blue{this large distribution shift}, the overall correspondence of team results remained consistent between the official CXR-LT challenge test set and the gold standard set (Fig.~\ref{fig:test-vs-gold}; $R^2 = 0.958$, $r=0.979$, $P=4.7\times10^{-6}$). \blue{Even the team rankings remained consistent with the exception of T3 outperforming T2 and T5 outperforming T4 on the gold standard test set. This is to be expected when considering that teams T2-T5 were separated by, at most, 0.005 mAP in the CXR-LT Test Phase.}

\subsection{\blue{Ranking stability analysis}}
\blue{All results presented thus far are point estimates that ignore the variability in predictive performance metrics -- for instance, how meaningful is it that T3 outperformed T4 by 0.001 mAP? We address this problem with a ranking stability analysis, whereby we perform 500 bootstrap samples of both the CXR-LT challenge test set and gold standard test set. Fig. \ref{fig:ranking-stability} depicts the percentage of simulated bootstrap ``trials" for which each team achieved a certain rank. On the CXR-LT challenge test set, ranking was generally stable, with the most common (``modal") simulated rank corresponding to the actual observed rank for all 9 teams. While T1 and T9 always, respectively, placed 1\textsuperscript{st} and 9\textsuperscript{th}, T2 and T3 were nearly interchangeable, with T2 placing 2\textsuperscript{nd} in 52\% of trials and T3 placing 2\textsuperscript{nd} in the remaining 48\%.}

\blue{In contrast, simulated team rankings were far more volatile when evaluated on the gold standard test set. For example, T1 – which considerably outperformed other teams in the official challenge – placed 1\textsuperscript{st} in under half of all simulated rankings, placing as low as 5\textsuperscript{th} in some instances. Additionally, T2 placed 1\textsuperscript{st} in 40\% of trials, perhaps supported by the fact that the cosine similarity between the predicted probabilities of T1 and T2 was 0.95, considerably higher than between T1 and any other team (Supplementary Fig. 3). Overall, this variability is to be expected given the small sample size of the gold standard set; even when sampling ``disease-heavy" studies to ensure that every finding was represented, certain tail classes were still only observed a handful of times (as few as 9), resulting in noisy estimates of performance. While improved label quality is valuable, this highlights the pervasive importance of sample size and justifies our choice for using the large-scale, text-mined labels for CXR-LT Test Phase evaluation. Despite this variability, however, the simulated modal rank corresponded to the original CXR-LT Test Phase rank in all instances but one: on the gold standard set, T4 placed 5\textsuperscript{th} most often (43\%) and T5 placed 4\textsuperscript{th} most often (52\%).}

\section{Discussion}

\subsection{Themes of successful solutions}

As outlined in Table \ref{tab:overview}, several salient patterns emerge among top-performing challenge solutions. \blue{In summary, many successful CXR-LT solutions leveraged
	\begin{itemize}[noitemsep, topsep=2pt, leftmargin=0.3in]
		\item Relatively high image resolution ($>$300 \texttimes\ 300)
		\item Modern CNNs like ConvNeXt and EfficientNetV2
		\item Large-scale domain-specific pretraining on CXR data
		\item Strong data augmentation and ensemble learning
		\item Loss re-weighting to amplify tail classes
		\item Multimodal learning via text-based label representations.
\end{itemize}}

\textbf{\blue{High image resolution.}} Compared to the vast majority of research efforts in natural image recognition, top-performing \blue{CXR-LT} solutions used relatively high image resolution \blue{-- usually greater than the standard 224 \texttimes\ 224 and} as high as 1024 \texttimes\ 1024. \blue{It is well-understood that increased image resolution improves visual recognition with deep neural networks \citep{touvron2019fixing,tan2019efficientnet,tan2021efficientnetv2}, particularly in medical applications such as radiology \citep{thambawita2021impact,sabottke2020effect}, where diagnosis can often rely on resolving faint or small abnormalities. \cite{haque2023effect} specifically found 1024 \texttimes\ 1024 (used by two top-performing teams) to be an optimal spatial resolution for MIMIC-CXR-JPG disease classification, observing that 2048 \texttimes\ 2048 resolution degraded performance.}

\textbf{\blue{Modern convolutional neural network backbones.}} Also, despite the recent popularity of Vision Transformers (ViTs) \citep{khan2022transformers}, all top-performing solutions used a convolutional neural network (CNN) as the image encoder. The most popular choice was ConvNeXt \citep{liu2022convnet}, followed by the EfficientNet \citep{tan2021efficientnetv2} and ResNet \citep{he2016deep} architecture families. \blue{This observation is echoed by a recent comprehensive benchmark of architectures on a wide array of visual recognition tasks \citep{goldblum2024battle}: ``Despite the recent attention paid to transformer-based architectures and self-supervised learning, high-performance convolutional networks pretrained via supervised learning outperform transformers on the majority of tasks..." The authors also found ConvNeXt, the most popular architecture among top CXR-LT performers, to outperform other backbones.}

\blue{\textbf{Large-scale pretraining.}} In this vein, all top-performing solutions utilized \blue{supervised} pretraining or transfer learning of some kind. While many used standard ImageNet-pretrained models (some leveraging the larger ImageNet21k), several teams performed additional domain-specific pretraining on publicly available, external CXR datasets \blue{such as NIH \blue{ChestX-Ray} \citep{wang2017chestx}, CheXpert \citep{irvin2019chexpert}, and PadChest \citep{bustos2020padchest}. Specifically, T1, T3, and T5 all performed multi-stage pretraining, whereby they fine-tuned an ImageNet-pretrained backbone on large external CXR datasets for disease classification. Such a two-stage pretraining scheme with (1) ``generalist" then (2) domain-specific pretraining has proven successful in prior works such as REMEDIS \citep{azizi2023robust} and \cite{reed2022self} in the context of self-supervised pretraining.}

\blue{\textbf{Ensemble learning and data augmentation.}} Many (7 out of 9) \blue{top} solutions involved \blue{ensemble learning, aggregating the outputs of multiple models} for improved generalization \blue{\citep{ganaie2022ensemble,fort2019deep}. Ensembles often benefit from diverse constituent models, and teams formed diverse ensembles in several unique ways: T2 and T5 formed ensembles across different model architectures, T2 and T7 formed ensembles across different image resolutions, and T3 formed an ensemble of head (common) class and tail (rare) class models. However, ensembling} was not \blue{\textit{strictly}} necessary for high performance, evidenced by the fact that the 1st- and 4th-placed teams utilized a single well-trained model. \blue{Additionally, all teams used image augmentation, another standard technique to boost generalization \citep{xu2023comprehensive}. Notably, T2 and T8 employed ``mosaic`` and CutMix augmentation, respectively, effectively blending input images and labels as a form of regularization.}

\blue{\textbf{Loss re-weighting.}} Owing to the challenging long-tailed nature of this problem, the top \blue{five} solutions all used loss \blue{re-weighting} in order to adequately model rare classes. \blue{From Table \ref{tab:lt-results}, one can see that the performance gap between T1-T5 and the remaining four teams is most apparent on the uncommon medium and tail classes. Specifically, T1 used a weighted asymmetric loss \citep{ridnik2021asymmetric} specifically designed for imbalanced multi-label classification, T2 used a class-weighted focal loss \citep{lin2017focal}, and T4 used a novel robust asymmetric loss that includes an additional regularization term to a weighted focal loss. It should also be noted that teams addressed the long-tailed distribution in ways other than loss re-weighting, such as mixture of experts (T3 and T9) and synthetic data generation of tail classes (T6).}

\blue{\textbf{Multimodal vision-language learning.} Multimodal vision-language learning has recently become a popular approach in deep learning for radiology, typically as a means of pretraining on paired CXR imaging and free-text radiology reports \citep{chen2019deep,yan2022clinical,delbrouck2022vilmedic,moon2022multi,li2024llava,moor2023med}. While this challenge did not use radiology report data, three top-performing teams found success by directly representing the disease label information as \textit{text}. For example, T1 used the ML-Decoder \citep{ridnik2023ml} classification head, which considers labels as text ``queries" that participate in cross-attention with image features. T5, on the other hand, used a Transformer pretrained on large amounts of clinical text \citep{pubmedbert,lehman2023clinical} to directly encode disease labels. For example, encoding the text ``Pleural Effusion" rather than a one-hot label enables the model to embed this information in a semantically meaningful way in relation to the language it has already encountered during pretraining on medical text.}

\subsection{Limitations and future work}

In addition to the common themes of successful solutions outlined above, it should be emphasized that the unique and often novel aspects of each team's solution also contributed to their success. For example, \cite{Kim_2023_ICCVa} leveraged a cross-view Transformer to aggregate information across radiographic views; \cite{Park_2023_ICCV} proposed a novel robust asymmetric loss (RAL), with additional experiments demonstrating improved performance on other long-tailed medical image classification tasks; \cite{verma2023can} leveraged a vision-language foundation model, RoentGen \citep{chambon2022roentgen}, to synthesize ``tail" class examples to combat the long-tailed problem; \cite{hong2023bag} took a vision-language approach leveraging Transformers pretrained on clinical text in order to learn rich representations of the multi-label disease information. One promising observation is that many teams reached very similar overall performance -- measured by Test Phase mAP -- with dramatically different methods. This suggests that the solutions from our participants may represent \textit{orthogonal} contributions that, when combined, prove greater than the sum of their parts. Future work might unify the insights learned from top-performing solutions into a single long-tailed, multi-label medical image classification framework.

\blue{This study could also be strengthened by examining bias both in the data and the models put forth by top-performing teams. First, MIMIC-CXR was collected at a single institution -- a major teaching hospital of Harvard Medical School -- which may only represent the specific demographics of this setting. Second, prior work has shown that deep neural networks trained on single-institution CXR datasets often display disparities in predictive performance based on factors like race and sex \citep{seyyed2021chexclusion}. \cite{seyyed2021chexclusion} observed this effect on MIMIC-CXR-JPG specifically and noted that training on larger, multi-institutional datasets mitigated these disparities. It would be interesting to evaluate whether teams that pretrained on additional external CXR data achieved not only improved predictive performance but also improved group fairness. Toward this end, the next iteration of CXR-LT will augment the gold standard test set with manually annotated data from another institution. Future work may also address subgroup fairness as an additional dimension of desired model behavior alongside predictive performance. }

Regarding the data contributions of this work, we acknowledge that the CXR-LT dataset bears the same pitfalls as many other publicly available CXR benchmarks with automatically text-mined labels, namely label noise \citep{abdalla2023hurdles}. We attempted to rectify this by manually annotating radiology reports to obtain a gold standard test set for additional evaluation. While this improved the label quality, this approach is limited in that it can only, at best, confirm the opinion of the individual radiologist writing the report. Even for highly trained experts, diagnosis from CXR is difficult and complex, leading to high inter-reader variability \citep{hopstaken2004inter,sakurada2012inter}. Ideally, a true ``gold standard" dataset would consist of consensus labeling from multiple radiologists' interpretations. However, this is of course prohibitively expensive and time-consuming \citep{zhou2021review} given the volume of labeled data required to train deep learning models and is the primary motivation for automatic disease labeling in the first place.

Though many diagnostic exams are long-tailed, most publicly available medical imaging datasets only include labels for a few common findings. The CXR-LT dataset thus represents a major contribution to due its large scale ($>$375,000 images), multi-label nature, and long-tailed label distribution of 26 clinical findings. However, a select few large-scale, long-tailed medical imaging datasets exist, such as HyperKvasir \citep{borgli2020hyperkvasir} -- containing $>$10,000 endoscopic images labeled with 23 findings -- and PadChest \citep{bustos2020padchest} -- containing $>$160,000 CXRs labeled with 174 findings.

While CXR-LT and PadChest represent meaningful contributions to long-tailed learning from CXR, it should be noted that the ``true" long tail of all clinical findings is at least an order of magnitude longer than any current publicly available dataset can offer. For example, Radiology Gamuts Ontology\footnote{\url{http://www.gamuts.net/about.php}} \citep{budovec2014informatics} documents 4,691 unique radiological image findings. Thus, one way to enhance the CXR-LT dataset might be to include an even wider variety of automatically text-mined findings to mimic the extremely long tail of real-world CXR. However, this approach too has its own limitations. Even if we could construct a dataset with labels for up to 1,000 clinical findings, ranging from common and well-studied to exceedingly rare, and train a model on this long-tailed data, what happens when a new finding is encountered? One might argue that the only way to tackle the \textit{true} long-tailed distribution of imaging findings is to develop a model that can adaptively generalize to previously unseen diseases. Future iterations of the CXR-LT challenge \blue{will} consider this problem through the lens of zero-shot classification: can participants train a model to accurately detect a clinical finding that the model has \textit{not been trained on}?

\subsection{The future of long-tailed, multi-label learning}

If zero-shot disease classification is the ultimate step toward clinically viable long-tailed medical image classification, then vision-language foundation models provide a very promising path forward. Several top-performing CXR-LT teams found success by encoding the label information as text, allowing for rich representation learning of the disease labels and their correlations. This allowed \cite{Kim_2023_ICCVa} to better handle the long-tailed, multi-label distribution via the text- and attention-based ML-Decoder classifier \citep{ridnik2023ml} and \cite{Seo_2023_ICCV} to exploit correlations between labels via graph learning of text-based label representations via ML-GCN \citep{chen2019multi}. While, for example, the approach of \cite{hong2023bag} did not earn 1\textsuperscript{st} place in this \blue{challenge}, it would almost certainly prove the most useful when encountering a previously unseen disease finding, a practical scenario in real-world clinical deployment. Since \cite{hong2023bag} leverage Transformer encoders pretrained on large collections of clinical text including PubMedBERT \citep{pubmedbert} and Clinical-T5 \citep{lehman2023clinical}, the model retains a rough \textit{semantic} understanding of biomedical concepts via the textual representation of disease information. This in turn allows natural generalization to new findings by relating the text \blue{``prompt"} of the potential new disease finding to relevant concepts encountered during pretraining, which is expressly not possible with standard unimodal deep learning methods.

Recent studies have shown that vision-language modeling with encoders pretrained on large collections of medical image and text data enable zero-shot disease classification, in some cases nearly reaching the performance of fully supervised approaches \citep{hayat2021multi,tiu2022expert,mishra2023improving,zhang2023knowledge,li2024llava,moor2023med}. \blue{For example, CheXzero \citep{tiu2022expert} performed Contrastive Language-Image Pretraining (CLIP) \citep{radford2021learning} on paired CXR images and radiology reports, where the model was trained to properly match reports with their CXRs. At test time, this model could then encode text ``prompts" of ``$<$Pathology$>$" and ``No $<$Pathology$>$", after which the model could determine which prompt best matched the given CXR, effectively performing zero-shot disease classification on par with board-certified radiologists. Critically, this ability to flexibly encode \textit{any} new disease information would enable zero-shot learning of tail classes for which it might otherwise be difficult or impossible to obtain traditional labels.} Further, such a multimodal approach can be readily combined with many of the methods employed by top-performing CXR-LT teams, such as loss re-weighting, heavy data augmentation, and ensemble learning. This would allow for a computer-aided diagnosis system capable of adaptively generalizing to unseen findings, thus capturing the true long tail of all imaging findings.

\section{Conclusion}

In summary, we curated and publicly released a large-scale dataset of over 375,000 CXR images for long-tailed, multi-label disease classification. We then hosted an open challenge, CXR-LT, to engage with the research community on this important task. We compiled and synthesized common threads through the most successful challenge solutions, providing practical recommendations for long-tailed, multi-label medical image classification. Lastly, we identify a path forward toward tackling the ``true" long tail of all imaging findings via multimodal vision-language foundation models capable of zero-shot generalization to unseen diseases, which future iterations of the CXR-LT challenge will address.

\section*{Declaration of competing interest}

\blue{R.M.S has received royalties for patent or software licenses from iCAD, Philips, PingAn, ScanMed, Translation Holdings, and MGB as well as research support form CRADA with PingAn.} The \blue{remaining} authors declare that they have no known competing financial interests or personal relationships that could have appeared to influence the work reported in this paper.

\section*{Acknowledgments}

This work was supported by the National Library of Medicine [grant number R01LM014306], the NSF [grant numbers 2145640, IIS-2212176], the Amazon Research Award, and the Artificial Intelligence Journal. It was also supported by the NIH Intramural Research Program, National Library of Medicine and Clinical Center. 
The authors would like to thank Alistair Johnson and the PhysioNet team for helping to publicize the \blue{challenge} and host the data.

\bibliographystyle{model2-names.bst} 
\bibliography{ref}

\renewcommand{\figurename}{Supplementary Figure}
\renewcommand{\tablename}{Supplementary Table}
\setcounter{figure}{0}
\setcounter{table}{0}

\clearpage

\begin{figure*}[!ht]
	\centering
	\includegraphics[scale=0.65]{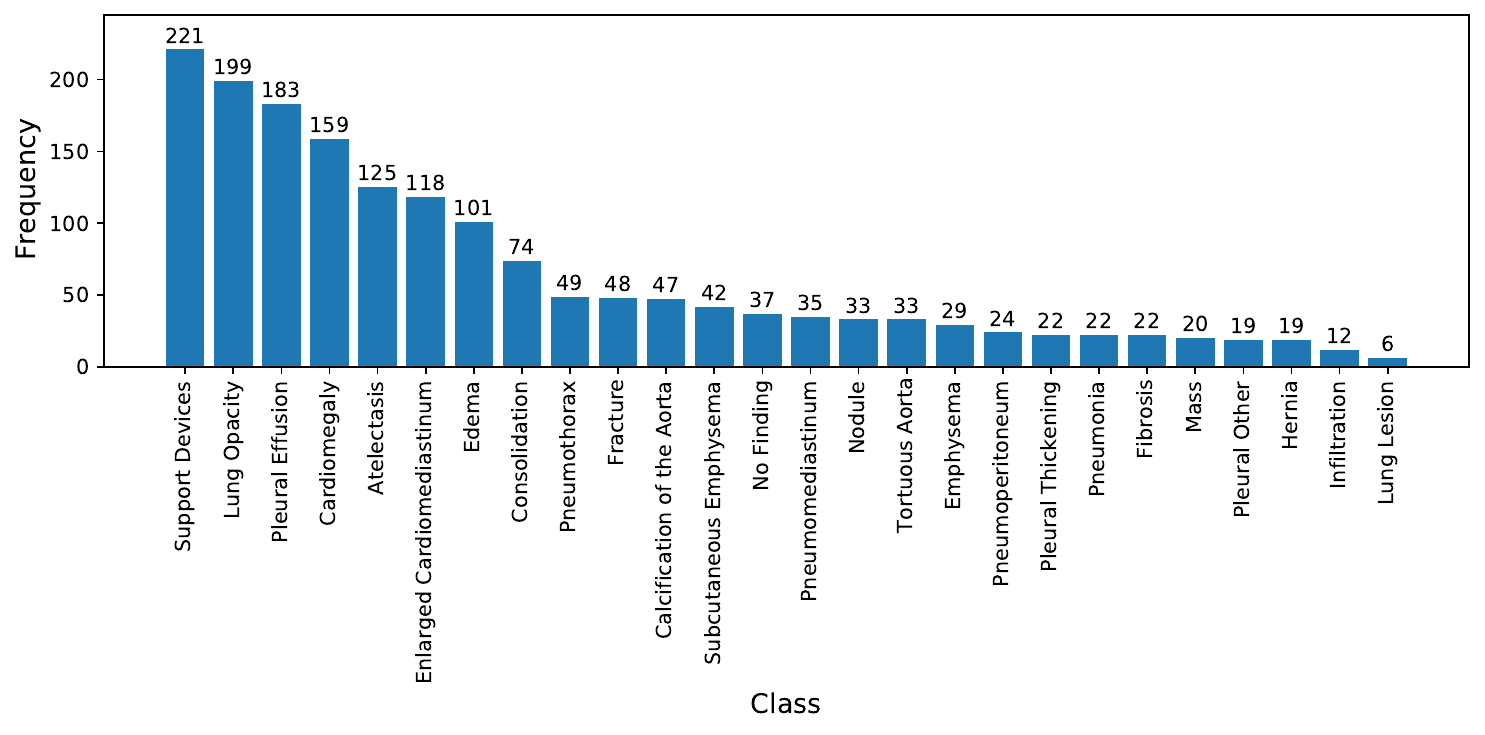}
	\caption{Long-tailed distribution of the CXR-LT gold standard test set.}
	\label{fig:enter-label}
\end{figure*}

\clearpage

\begin{figure*}[!ht]
	\centering
	\includegraphics[scale=0.575]{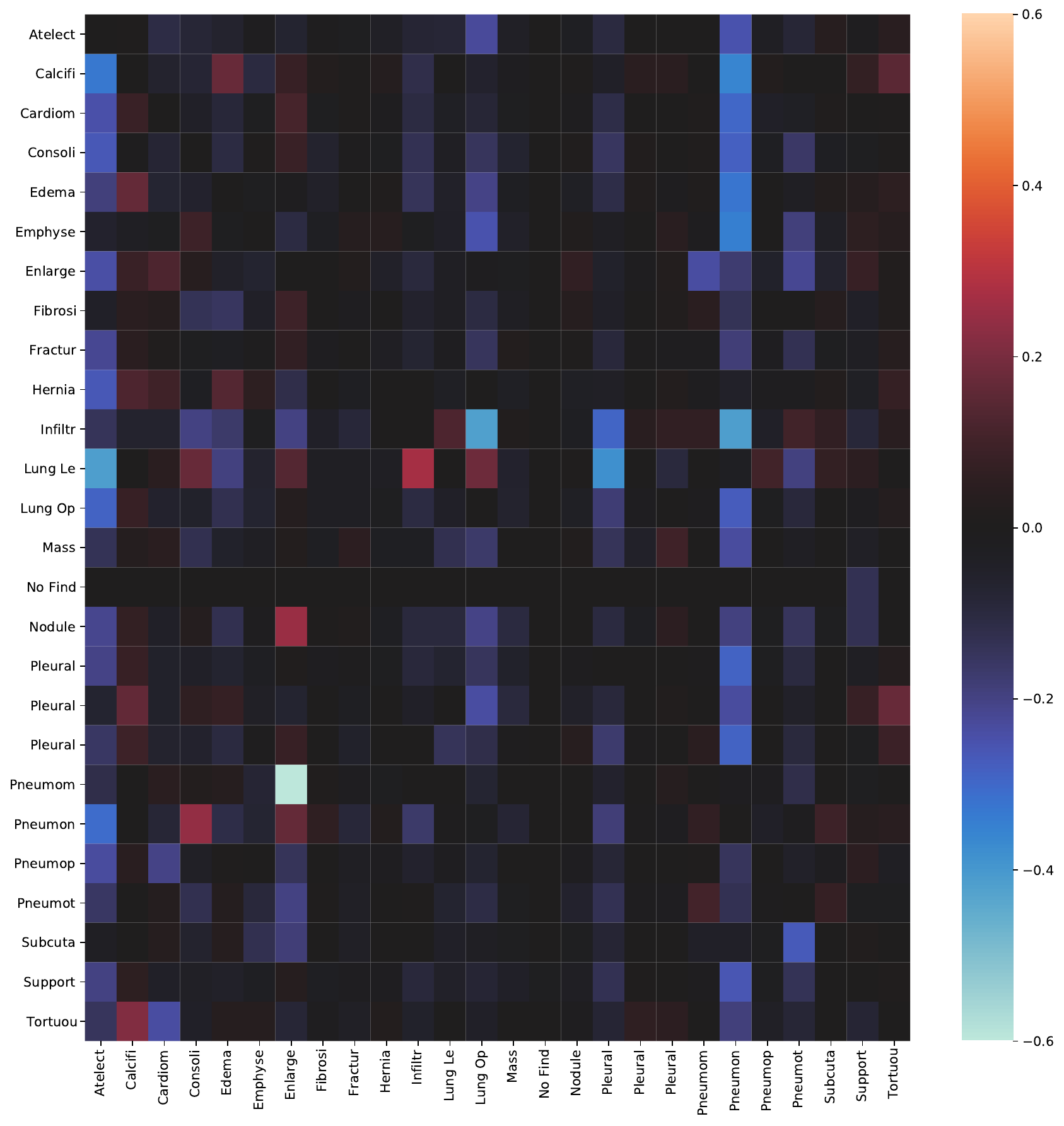}
	\caption{Heatmap displaying the difference in class co-occurrence tendencies between automatically text-mined and manually human-annotated labels. Each entry depicts the difference in conditional probability of the x-axis finding given the y-axis finding between the gold standard human labels and automatically text-mined labels.}
	\label{fig:co-occurrence}
\end{figure*}

\clearpage

\begin{figure}[ht]
	\centering
	\includegraphics[scale=0.8]{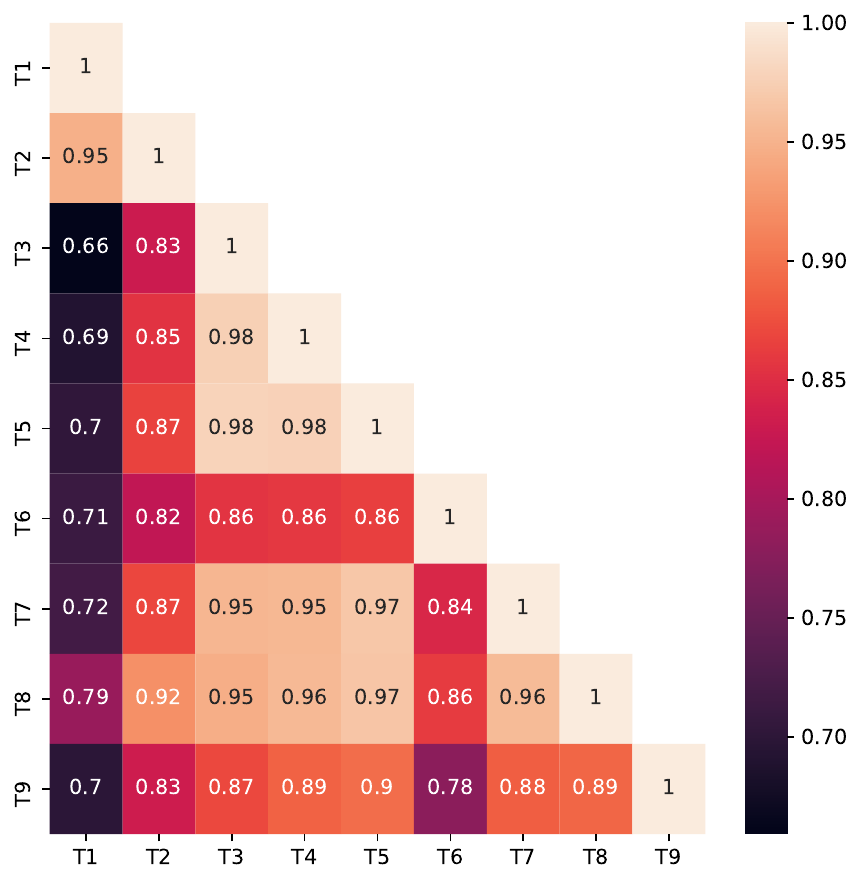}
	\caption{Heatmap displaying cosine similarities of predicted probabilities between top-performing teams on the CXR-LT challenge test set.}
	\label{fig:similarity}
\end{figure}

\clearpage

\begin{table*}[htbp]
	\centering
	\caption{\label{tab:auc-results}Final Test Phase results of the CXR-LT 2023 competition. Presented is area under the receiver operating characteristic curve (AUROC) of each team's final model on all 26 classes evaluated on the test set. The best AUROC for a given class is highlighted in bold.}
	{\small
		\begin{tabular}{lcccccccccc}
			\toprule
			& T1 & T2 & T3 & T4 & T5 & T6 & T7 & T8 & T9 \\ \midrule
			Atelectasis & \textbf{0.838} & 0.828 & 0.828 & 0.827 & 0.826 & 0.828 & 0.825 & 0.819 & 0.790 \\
			Calcification of the Aorta & \textbf{0.920} & 0.896 & 0.891 & 0.893 & 0.892 & 0.871 & 0.875 & 0.871 & 0.857 \\
			Cardiomegaly & 0.815 & 0.810 & 0.809 & 0.807 & 0.810 & 0.809 & \textbf{0.818} & 0.803 & 0.770 \\
			Consolidation & \textbf{0.794} & 0.787 & 0.789 & 0.783 & 0.784 & 0.789 & 0.781 & 0.779 & 0.733 \\
			Edema & \textbf{0.858} & 0.855 & 0.857 & 0.857 & 0.854 & 0.855 & 0.853 & 0.851 & 0.830 \\
			Emphysema & \textbf{0.916} & 0.903 & 0.911 & 0.909 & 0.899 & 0.907 & 0.893 & 0.895 & 0.871 \\
			Enlarged Cardiomediastinum & \textbf{0.619} & 0.617 & 0.615 & 0.617 & 0.615 & 0.615 & 0.61 & 0.605 & 0.557 \\
			Fibrosis & \textbf{0.922} & 0.916 & 0.921 & 0.919 & 0.907 & 0.912 & 0.900 & 0.901 & 0.838 \\
			Fracture & \textbf{0.852} & 0.795 & 0.799 & 0.814 & 0.788 & 0.799 & 0.756 & 0.774 & 0.722 \\
			Hernia & 0.914 & \textbf{0.916} & 0.910 & 0.909 & 0.909 & 0.905 & 0.899 & 0.891 & 0.853 \\
			Infiltration & 0.630 & 0.615 & 0.627 & \textbf{0.634} & 0.615 & 0.621 & 0.609 & 0.611 & 0.58 \\
			Lung Lesion & \textbf{0.802} & 0.791 & 0.790 & 0.791 & 0.795 & 0.769 & 0.760 & 0.769 & 0.717 \\
			Lung Opacity & \textbf{0.800} & 0.784 & 0.786 & 0.783 & 0.781 & 0.786 & 0.780 & 0.774 & 0.737 \\
			Mass & \textbf{0.821} & 0.811 & 0.811 & 0.800 & 0.807 & 0.814 & 0.803 & 0.789 & 0.727 \\
			Nodule & \textbf{0.846} & 0.806 & 0.809 & 0.804 & 0.796 & 0.803 & 0.769 & 0.781 & 0.737 \\
			Pleural Effusion & \textbf{0.930} & 0.921 & 0.922 & 0.923 & 0.922 & 0.909 & 0.917 & 0.917 & 0.901 \\
			Pleural Other & \textbf{0.910} & 0.878 & 0.876 & 0.882 & 0.890 & 0.857 & 0.886 & 0.872 & 0.751 \\
			Pleural Thickening & \textbf{0.888} & 0.835 & 0.843 & 0.842 & 0.847 & 0.793 & 0.848 & 0.819 & 0.763 \\
			Pneumomediastinum & 0.918 & 0.934 & 0.933 & 0.933 & \textbf{0.939} & 0.930 & 0.917 & 0.911 & 0.846 \\
			Pneumonia & 0.657 & 0.650 & 0.657 & \textbf{0.658} & 0.651 & 0.654 & 0.641 & 0.645 & 0.609 \\
			Pneumoperitoneum & 0.908 & 0.901 & 0.900 & 0.900 & \textbf{0.913} & 0.881 & 0.893 & 0.862 & 0.815 \\
			Pneumothorax & \textbf{0.886} & 0.876 & 0.878 & 0.880 & 0.873 & 0.875 & 0.858 & 0.858 & 0.833 \\
			Subcutaneous Emphysema & \textbf{0.990} & 0.978 & 0.987 & 0.986 & 0.982 & 0.982 & 0.98 & 0.975 & 0.965 \\
			Support Devices & \textbf{0.956} & 0.946 & 0.952 & 0.952 & 0.947 & 0.948 & 0.938 & 0.942 & 0.934 \\
			Tortuous Aorta & \textbf{0.841} & 0.834 & 0.821 & 0.825 & 0.829 & 0.817 & 0.824 & 0.814 & 0.777 \\
			No Finding & \textbf{0.861} & 0.855 & 0.858 & 0.855 & 0.855 & 0.845 & 0.851 & 0.849 & 0.821 \\ \midrule
			Mean & \textbf{0.850} & 0.836 & 0.838 & 0.838 & 0.836 & 0.830 & 0.826 & 0.822 & 0.782 \\ \bottomrule
		\end{tabular}
	}
\end{table*}

\clearpage

\begin{table*}[!ht]
	\caption{\label{tab:gold-auc}Gold standard test set results from CXR-LT 2023 participants. Presented is area under the receiver operating characteristic (AUROC) of each team's final model on all 26 classes evaluated on our human-annotated gold standard test set. The best AUROC for a given class is highlighted in bold.}
	\centering
	{\small
		\begin{tabular}{lcccccccccc}
			\toprule
			& T1 & T2 & T3 & T4 & T5 & T6 & T7 & T8 & T9 \\ \midrule
			Atelectasis & 0.702 & 0.698 & 0.706 & 0.694 & \textbf{0.713} & 0.699 & 0.693 & 0.692 & 0.687 \\
			Calcification of the Aorta & \textbf{0.928} & 0.918 & 0.926 & 0.922 & 0.911 & 0.890 & 0.855 & 0.895 & 0.876 \\
			Cardiomegaly & 0.798 & 0.811 & 0.812 & 0.807 & 0.817 & \textbf{0.824} & 0.821 & 0.808 & 0.764 \\
			Consolidation & 0.746 & 0.760 & 0.766 & 0.762 & 0.752 & \textbf{0.766} & 0.748 & 0.728 & 0.719 \\
			Edema & \textbf{0.839} & 0.837 & 0.833 & 0.822 & 0.833 & 0.831 & 0.820 & 0.828 & 0.809 \\
			Emphysema & 0.863 & \textbf{0.881} & 0.876 & 0.864 & 0.876 & 0.872 & 0.877 & 0.853 & 0.854 \\
			Enlarged Cardiomediastinum & 0.574 & \textbf{0.588} & 0.583 & 0.561 & 0.580 & 0.577 & 0.564 & 0.542 & 0.575 \\
			Fibrosis & 0.884 & 0.858 & 0.880 & 0.868 & \textbf{0.886} & 0.867 & 0.841 & 0.852 & 0.872 \\
			Fracture & \textbf{0.887} & 0.823 & 0.856 & 0.837 & 0.828 & 0.841 & 0.808 & 0.794 & 0.809 \\
			Hernia & 0.938 & 0.928 & 0.929 & 0.944 & 0.925 & 0.940 & \textbf{0.945} & 0.889 & 0.909 \\
			Infiltration & \textbf{0.674} & 0.571 & 0.578 & 0.598 & 0.565 & 0.571 & 0.575 & 0.584 & 0.633 \\
			Lung Lesion & 0.651 & 0.704 & 0.674 & \textbf{0.712} & 0.650 & 0.699 & 0.637 & 0.654 & 0.699 \\
			Lung Opacity & 0.688 & 0.690 & \textbf{0.700} & 0.676 & 0.699 & 0.686 & 0.693 & 0.698 & 0.667 \\
			Mass & 0.828 & 0.861 & 0.851 & 0.842 & 0.841 & \textbf{0.884} & 0.829 & 0.783 & 0.733 \\
			Nodule & \textbf{0.801} & 0.706 & 0.713 & 0.737 & 0.728 & 0.791 & 0.734 & 0.724 & 0.730 \\
			Pleural Effusion & \textbf{0.880} & 0.860 & 0.865 & 0.870 & 0.864 & 0.847 & 0.853 & 0.861 & 0.851 \\
			Pleural Other & \textbf{0.893} & 0.833 & 0.858 & 0.852 & 0.890 & 0.760 & 0.836 & 0.869 & 0.767 \\
			Pleural Thickening & \textbf{0.826} & 0.767 & 0.759 & 0.751 & 0.777 & 0.785 & 0.784 & 0.773 & 0.72 \\
			Pneumomediastinum & 0.953 & \textbf{0.982} & 0.974 & 0.973 & 0.98 & 0.972 & 0.982 & 0.967 & 0.849 \\
			Pneumonia & 0.631 & \textbf{0.677} & 0.663 & 0.650 & 0.659 & 0.653 & 0.641 & 0.656 & 0.615 \\
			Pneumoperitoneum & 0.882 & 0.881 & 0.903 & 0.905 & \textbf{0.907} & 0.888 & 0.884 & 0.890 & 0.824 \\
			Pneumothorax & 0.942 & 0.934 & \textbf{0.944} & 0.943 & 0.926 & 0.940 & 0.905 & 0.924 & 0.906 \\
			Subcutaneous Emphysema & 0.986 & 0.973 & \textbf{0.988} & 0.988 & 0.987 & 0.978 & 0.978 & 0.973 & 0.985 \\
			Support Devices & 0.948 & 0.941 & 0.946 & \textbf{0.950} & 0.948 & 0.941 & 0.918 & 0.939 & 0.934 \\
			Tortuous Aorta & \textbf{0.861} & 0.831 & 0.822 & 0.828 & 0.833 & 0.829 & 0.811 & 0.824 & 0.806 \\
			No Finding & 0.944 & \textbf{0.957} & 0.953 & 0.943 & 0.952 & 0.95 & 0.952 & 0.944 & 0.938 \\ \midrule
			Mean & \textbf{0.829} & 0.818 & 0.821 & 0.819 & 0.82 & 0.819 & 0.807 & 0.805 & 0.790 \\
			\bottomrule
		\end{tabular}
	}
\end{table*}

\clearpage

\begin{table}[ht]
	\centering
	\caption{Comparison of predictive performance (average precision) with automated text-mined labeling vs. manual human annotation of the 406 radiology reports in our gold standard test set.}
	\begin{tabular}{@{}lccr@{}}
		\toprule
		& Automated & Human & \makecell[c]{$\Delta$} \\ \midrule
		Atelectasis & 0.439 & 0.440 & $+$0\% \\
		Calcification of the Aorta & 0.798 & 0.627 & $-$21\% \\
		Cardiomegaly & 0.729 & 0.582 & $-$20\% \\
		Consolidation & 0.522 & 0.470 & $-$10\% \\
		Edema & 0.585 & 0.342 & $-$42\% \\
		Emphysema & 0.730 & 0.651 & $-$11\% \\
		Enlarged Cardiomediastinum & 0.467 & 0.338 & $-$28\% \\
		Fibrosis & 0.820 & 0.773 & $-$6\% \\
		Fracture & 0.183 & 0.060 & $-$67\% \\
		Hernia & 0.495 & 0.534 & $+$8\% \\
		Infiltration & 0.277 & 0.043 & $-$84\% \\
		Lung Lesion & 0.407 & 0.115 & $-$72\% \\
		Lung Opacity & 0.728 & 0.467 & $-$36\% \\
		Mass & 0.915 & 0.830 & $-$9\% \\
		No Finding & 0.272 & 0.229 & $-$16\% \\
		Nodule & 0.261 & 0.313 & $+$20\% \\
		Pleural Effusion & 0.795 & 0.714 & $-$10\% \\
		Pleural Other & 0.956 & 0.950 & $-$1\% \\
		Pleural Thickening & 0.238 & 0.321 & $+$35\% \\
		Pneumomediastinum & 0.822 & 0.800 & $-$3\% \\
		Pneumonia & 0.442 & 0.452 & $+$2\% \\
		Pneumoperitoneum & 0.243 & 0.235 & $-$3\% \\
		Pneumothorax & 0.643 & 0.724 & $+$13\% \\
		Subcutaneous Emphysema & 0.387 & 0.360 & $-$7\% \\
		Support Devices & 0.405 & 0.613 & $+$51\% \\
		Tortuous Aorta & 0.870 & 0.832 & $-$4\% \\ \bottomrule
	\end{tabular}
\end{table}

\end{document}